# Multi-Space Evolutionary Search for Large-Scale Optimization

Liang Feng, Qingxia Shang, Yaqing Hou, Kay Chen Tan and Yew-Soon Ong

*Abstract:* In recent years, to improve the evolutionary algorithms used to solve optimization problems involving a large number of decision variables, many attempts have been made to simplify the problem solution space of a given problem for the evolutionary search. In the literature, the existing approaches can generally be categorized as decomposition-based methods and dimension-reduction-based methods. The former decomposes a large-scale problem into several smaller subproblems, while the latter transforms the original high-dimensional solution space into a low-dimensional space. However, it is worth noting that a given large-scale optimization problem may not always be decomposable, and it is also difficult to guarantee that the global optimum of the original problem is preserved in the reduced low-dimensional problem space. This paper thus proposes a new search paradigm, namely the multi-space evolutionary search, to enhance the existing evolutionary search methods for solving large-scale optimization problems. In contrast to existing approaches that perform an evolutionary search in a single search space, the proposed paradigm is designed to conduct a search in multiple solution spaces that are derived from the given problem, each possessing a unique landscape. The proposed paradigm makes no assumptions about the large-scale optimization problem of interest, such as that the problem is decomposable or that a certain relationship exists among the decision variables. To verify the efficacy of the proposed paradigm, comprehensive empirical studies in comparison to four state-of-the-art algorithms were conducted using the CEC2013 large-scale benchmark problems.

## I. Introduction

An evolutionary algorithm (EA) is a stochastic optimization search method that takes inspiration from the theory of natural biological evolution. It starts with a population of individuals that undergo reproduction, including crossover and mutation, to produce offspring. This procedure is executed iteratively and terminated when a predefined condition is satisfied. In contrast to traditional optimization approaches such as calculus-based and enumerative strategies, an EA contains flexible procedures and is robust to changing problem circumstances. In recent decades, EAs have attracted significant research attention, and have been successfully applied in many complex applications, such as scheduling in logistics [1], [2], image processing [3], [4], and the architecture optimization of deep neural networks [5], [6].

Today, because of the exponential growth of the volume of data in big data applications, large-scale optimization problems (i.e., optimization problems with a large number of decision variables) have become ubiquitous in the real world [7], [8], [9], [10], [11]. Because increasing the decision variables not only leads to an exponential increase in the problem solution space, but also results in the growth of the computational cost involved in the solution search and evaluation process, the performance of an EA decreases significantly for large-scale optimization problems [12], [13], [14], [15]. To improve the scalability of EA for solving problems with large-scale dimensionality, many research efforts have been conducted to simplify the search space of a large-scale optimization problem [16], [17], [18], [19], [20], [21]. According to a recent survey [22], the existing approaches can generally be categorized into two groups: decomposition-based approaches and dimension-reduction-based approaches. Specifically, decomposition-based methods follow the principle of divide-and-conquer and decompose a problem into several relatively small subcomponents that are optimized concurrently. Further, dimension-reduction-based methods reduce the number of decision variables by selecting a set of principal variables or transforming the high-dimensional space into a space with fewer dimensions. However, despite the successes enjoyed by these two classes of methods, it is worth noting here that because decomposition-based methods rely on the accurate detection of the relationships between decision variables, this type of method may fail when used for large-scale optimization problems that possess complex variable interactions, or are not even decomposable. Moreover, reducing the dimensionality of the search space could lose important information for optimization, and it is difficult to guarantee that the global optimum or high-quality solutions are preserved in the reduced solution space.

Evolutionary multitasking (EMT) is a recently emerging research topic in the field of evolutionary computation [23], [24], [25], [26]. In contrast to a traditional single-task evolutionary search, EMT conducts an evolutionary search on multiple tasks, each corresponding to a particular optimization problem. It aims to improve the convergence characteristics of an evolutionary search across multiple optimization problems at once by seamlessly transferring knowledge among the tasks. In the literature, the efficacy of EMT has been verified on sets of continuous, discrete tasks, and mixtures of continuous and combinatorial tasks [27], [28], [29], [30], [31]. Inspired by this, in the context of large-scale optimization, besides replacing the original problem space with a simplified space, the advantage of a simplified search space could also be obtained by configuring the constructed solution space as an auxiliary task of the original problem, under EMT. In this manner, as the solution space of the original problem serves as one task in EMT, there is no assumption and requirement for the relationships between decision variables, and the existence of global optimum or high-quality solutions is guaranteed.

Keeping the above in mind, this paper proposes a new evolutionary search paradigm, namely the multi-space evolutionary search, for solving large-scale optimization problems. In particular, for a given large-scale optimization problem, besides the original problem space, multiple simplified solution spaces are derived for the given problem, which possess unique landscapes. Further, instead of conducting an evolutionary search on the given problem space, evolutionary searches are concurrently performed on both the original and constructed simplified spaces of the given problem. By transferring useful traits while the search progresses online across different spaces, via EMT, an enhanced problem-solving process can be obtained. To evaluate the efficacy of the proposed paradigm for large-scale optimization, comprehensive empirical studies on the CEC2013 large-scale optimization benchmarks were conducted using four state-of-the-art representative methods for large-scale optimization, and the results were analyzed.

The reminder of this paper is organized as follows. Section II begins with a review of the literature on the existing decomposition- and dimension-reduction-based approaches for solving large-scale optimization problems. A brief introduction of EMT, which serves as the optimization engine of the proposed multi-space evolutionary search, is also presented in section II. Further, section III provides the details of the proposed multi-space evolutionary search for solving largescale optimization problems. Section IV discusses the comprehensive empirical studies that were conducted on the CEC2013 large-scale optimization benchmarks using four state-of-art algorithms for large-scale optimization. Finally, the concluding remarks of this paper are presented in section V.

## II. Preliminary

This section first presents a review of the literature on the existing decomposition- and dimension-reduction-based approaches for solving large-scale optimization problems. Next, a brief introduction of the EMT paradigm is provided.

*A. Existing Approaches for Simplifying Search Space of large-scale Optimization Problems*

According to recent surveys in the literature [8], [22], [32], the existing approaches to simplifying the search space of a given large-scale optimization problem can generally be categorized as decomposition-based approaches, and dimension-reduction-based methods. In particular, the decomposition-based approaches are also known as divide-and-conquer approaches in evolutionary computation and mainly involve cooperative coevolution (CC) algorithms, which decompose a given large-scale optimization problem into several smaller subproblems and then optimize each subproblem separately using different EAs. Generally, decomposition-based approaches consist of three major steps. First, by considering the structure of the underlying decision variable interactions, the original D-dimensional problem is exclusively divided into $N$ $d_i$-dimensional sub-problems, where $\sum_{i=1}^{N} d_i = D$. Next, each subproblem is solved by a particular EA. Finally, the $d$-dimensional solutions to these subproblems are merged to form the D-dimensional complete solution for the original problem. It is straightforward to see how the decomposition of the problem is essential to the performance of CC algorithms, and how an inappropriate decomposition of the decision variables may even lead to a deteriorated optimization performance [32], [33]. Particular examples in this category include strategies that randomly divide the variables into groups without taking the variable interaction into consideration [34], [35], [36], approaches that make use of evolutionary information to learn variable interdependency and then divide variables into groups [37], [38], and static decomposition methods that are performed before conducting an evolutionary search based on the detection of variable interaction [33], [39], [40], [41].

On the other hand, instead of decomposing the solution space of the given problem, a dimension-reduction-based approach attempts to create a new solution space with lower dimensionality from the original solution space. The evolutionary search is then performed on the newly created low dimension space, and the obtained solution is mapped back to the original space for evaluation. Generally, the existing approaches perform dimension reduction either by selecting a subset of the original decision variables or transforming the original solution space into a low-dimensional solution space. As can be observed, the preservation of important information for guiding the search toward high-quality solutions in the reduced solution space plays a key role in determining the performance of a dimension-reduction-based approach. Examples belonging to this class include the random matrix projection-based estimation of distribution algorithm [42], random embedding-based approach for large-scale optimization problems with low effective dimensions [43], and multi-agent system assisted embedding for large-scale optimization [44].

Although the above methods have shown good performances in solving large-scale optimization problems, there are two main drawbacks with these two categories of methods. First, because decomposition-based methods rely heavily on the accurate detection of decision variable interactions, these methods may fail on large-scale optimization problems with complex variable interactions or that are not decomposable. Second, although dimension reduction may not rely on variable interaction, it is difficult to guarantee that the global optimum or high-quality solutions are preserved in the reduced space. However, because a simplified solution space can provide useful information for efficient and effective problem solving, it is desirable to develop new search paradigms for large-scale optimization that can leverage the advantage of simplified solution spaces without the limitations discussed above.

*B. Evolutionary Multitasking*

Consider a situation where $K$ optimization tasks are to be performed. EMT has been defined in the literature as an optimization paradigm that solves multiple optimization tasks at the same time, with the aim of improving the problem-solving performance across tasks by seamlessly transferring knowledge between them [23]. In particular, as depicted in Fig.1, let $f_i: \mathcal{X}_i \rightarrow \mathbb{R}$ be a global optimization task on a compact subset $\mathcal{X}_i \in \mathbb{R}^{D_i}$, with objective $x_i^* = argmin_{x_i \in \mathcal{X}_i} f_i(x_i)$, The input of EMT is a set of optimization tasks **IS**: $\{f_1, \cdots, f_i, \cdots, f_K\}$, where $K$ denotes the number of tasks. Please note that each task $f_i$ may possess unique dimensionality $D_i$. The output of EMT is then given by the set of optimized solutions **OS**: $\{x_1^*, \cdots, x_i^*, \cdots, x_K^*\}$.

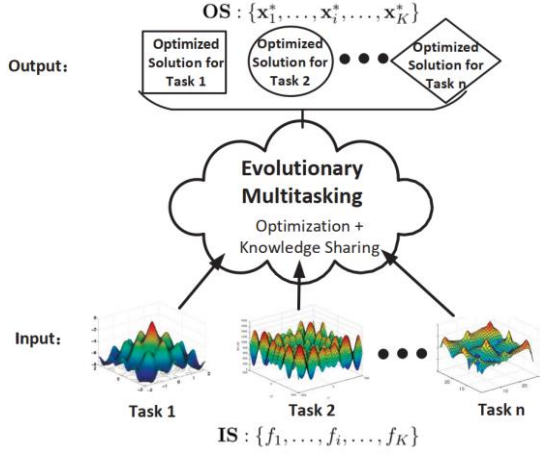

Fig. 1. Illustration of EMT.

In contrast to the traditional single-task optimization, EMT involves automatically exploiting and transferring the latent synergies between distinct (but possibly similar) optimization problems while the optimization progresses online, which could eventually lead to enhanced problem-solving on all the tasks. For large-scale optimization, if each task under EMT corresponds to a unique solution space for the given optimization problem, the useful traits found in different spaces could be transferred across these spaces via EMT, producing a more efficient and effective evolutionary search process. Inspired by this, a multi-space evolutionary search paradigm is proposed for large-scale optimization, which will be discussed in detail in the next section.

## III. Proposed Multi-space Evolutionary Search for Large-scale Optimization

This section presents the details of the proposed multi-space evolutionary search for large-scale optimization. In particular, the outline of the proposed paradigm is presented in Fig. 2. For a given problem of interest, besides the original problem space, a simplified problem space for the given problem is first created. Next, the mapping between these two problem spaces is learned, which will be used for knowledge transfer across spaces during the evolutionary search process via EMT. Further, by treating these two problem spaces as two tasks, evolutionary searches can be conducted on the tasks concurrently. As can be observed in the figure, knowledge transfer will be performed across tasks while the evolutionary search progresses online (see the green rectangle in Fig. 2). In this way, the useful traits found in the simplified problem space can be leveraged to facilitate the search in the original space, while the high-quality solutions found in the original problem space may also guide the search direction in the simplified problem space toward promising areas. Furthermore, to explore the usefulness of diverse auxiliary tasks, the simplified problem space will be re-constructed periodically using the solutions found during the evolutionary search process (see the yellow rectangle in Fig. 2). Finally, the EMT process is terminated when certain stopping criteria are satisfied.

The following sections present details on the *Construction of the simplified problem space*, *Learning of Mapping across Problem Spaces*, *Knowledge Transfer across Problem Spaces*, and *Reconstruction of the simplified problem space*.

### A. Construction of The Simplified Problem Space

Because the simplified problem space serves as an auxiliary task of a given problem of interest, there are generally no particular constraints on the construction of the simplified space. Therefore, the existing approaches proposed in the literature, such as random embedding [43], dimension reduction [45], or even search space decomposition [34], [38] could be employed for constructing the space.

In this paper, for simplicity, the popular and well-known dimension reduction approach principal component analysis (PCA) [46] is considered for constructing a simplified problem space, $\mathbf{P}_s$, in the proposed multi-space evolutionary search paradigm. In particular, to generate initial population $\mathbf{PoP}_s$ of the evolutionary search in $\mathbf{P}_s$, initial population $\mathbf{PoP}$ is first sampled in original problem space $\mathbf{P}$, which is routine [47]. Next, the obtained $\mathbf{PoP}$ in $\mathbf{P}$ will undergo PCA with dimension $d_s$ to generate $\mathbf{PoP}_s$ for the evolutionary search in $\mathbf{P}_s$.

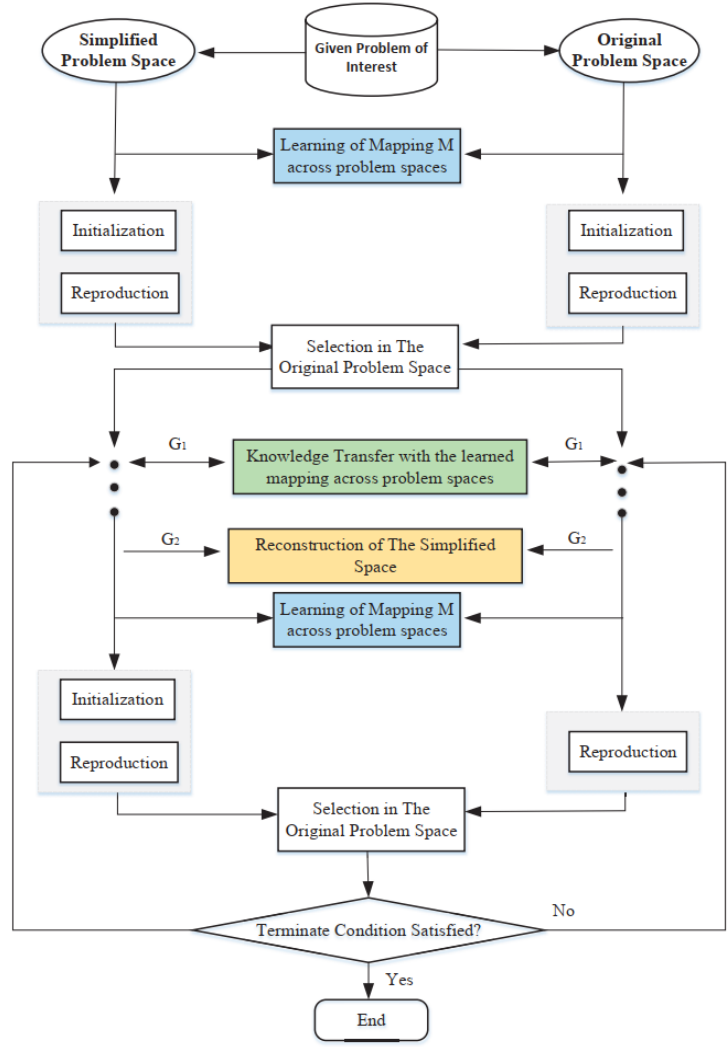

Fig. 2. Workflow of the proposed multi-space evolutionary search for large-scale optimization.

*B. Learning of Mapping across Problem Spaces*

Once the simplified problem space has been constructed, the mappings across simplified problem space $P_s$ and original problem space $P$ have to be learned, to allow the useful traits found in each space to be transferred across spaces toward efficient and effective problem solving for large-scale optimization.

In this paper, the mappings across $P_s$ and $P$ are learned using labeled data from each space via supervised learning. In particular, as discussed in section III-A, $PoP_s$ is generated by performing a PCA of $PoP$. Therefore, each solution in $PoP_s$ has a unique corresponding solution in $PoP$. This correspondence thus provides the label information to connect space $P_s$ and $P$. Taking this cue, by configuring $T$ and $S$ as $PoP$ and $PoP_s$, respectively, the mapping $M_{P_s \to P}: \mathcal{R}^{d_s} \to \mathcal{R}^d$ ($d$ is the dimension of the original problem space) from the simplified space $P_s$ to the original problem space $P$ can then be obtained by minimizing the squared reconstruction loss, which is given by the following:

$$\mathcal{L}_{sq}(M) = \frac{1}{2N} \sum_{i=1}^{N} \|p_i - M \times q_i\|^2 \quad (1)$$

where $N$ denotes the number of solutions in $S$ and $T$. $q_i$ is the solution in $S$, and $p_i$ gives the solution in $T$ which corresponds to $q_i$.

Further, to simplify the notation, it is assumed that a constant feature is added to the input, that is, $p_i = [p_i; 1]$ and $q_i = [q_i; 1]$, and an appropriate bias is incorporated within the mapping, $M = [M, b]$. The loss in Eq. 1 is then reduced to the matrix form:

$$\mathcal{L}_{sq}(M) = \frac{1}{2N} tr[(T - M \times S)^T (T - M \times S)] \quad (2)$$

Where $tr(\cdot)$ and $T$ denote the trace operation and transpose operation of a matrix, respectively. The solution of Eq. 2 can be expressed as the well-known closed-form solution for ordinary least squares [48], which is given by the following:

$$M = (T \times S^T)(S \times S^T)^{-1} \quad (3)$$

Finally, it is straightforward to see that the mapping $M_{P \to P_s}$ from space $\mathbf{P}$ to $\mathbf{P}_s$ can also be learned via Eq. 3 by configuring $T$ and $S$ as $\mathbf{PoP}_s$ and $\mathbf{PoP}$, respectively.

*C. Knowledge Transfer across Problem Spaces*

With the learned mapping $M_{P_s \to P}$ and $M_{P \to P_s}$ across the simplified and original problem spaces, a knowledge transfer across these two spaces can be easily conducted by the simple operation of matrix multiplication. In particular, for a knowledge transfer from $\mathbf{P}_s$ to $\mathbf{P}$, suppose this process occurs at generation $G_1$. First, the Q best solutions are selected in terms of the fitness values from the population of the simplified problem space, dented by $S_s$, which is a $d_s \times Q$ matrix. Next, the transferred solutions, $TS_{P_s \to P}$, are obtained by $M_{P_s \to P} \times S_s$. Finally, the solutions in $TS_{P_s \to P}$ are injected into the population of the original problem space to undergo natural selection for the next generation.

On the other hand, at generation $G_1$, knowledge transfer also occurs from the original problem space to the simplified problem space. In particular, the $P$ best solutions in terms of fitness value are first selected from the population of the original problem space, which are labeled as $S'_s$ and are a $d \times P$ matrix. Subsequently, the transferred solutions $TS_{P \to P_s}$ can be obtained by $M_{P \to P_s} \times S'_s$. Further, the solutions in $TS_{P \to P_s}$ are inserted into the population of the simplified problem space with natural selection.

Moreover, after the knowledge transfer process, the updated population of the simplified problem space is further transformed backed to the original problem space, and archived in $A_s$. The repeated solutions in $A_s$ are removed. As can be observed, $A_s$ preserves

---

**Algorithm 1:** Pseudo code of the population re-initialization in $\mathbf{P}_s$

**Input:** $\mathbf{PoP}_s$: the population in the simplified problem space before the reconstruction; $A_s$: Archive in the original problem space

**Output:** $\mathbf{PoP}_s$: the re-initialized population in the new simplified problem space

1: **Begin**
2: **Transform** $\mathbf{PoP}_s$ back to the original problem space using $M_{P_s \to P}$, denoted by $\mathbf{PoP}'_s$;
3: **Perform** the reconstruction of the simplified problem space using PCA with $A_s$;
4: **Learn** the new $M_{P_s \to P}$ and $M_{P \to P_s}$ across problem spaces $\mathbf{P}$ and $\mathbf{P}_s$ as discussed in section III-B.
5: **Re-initialize** the population in the new simplified space by $PoP_s = M_{P \to P_s} \times PoP'_s$;
6: **End**

---

search traces in the present simplified problem space, which will be used for the reconstruction of a new simplified space. That is discussed in detail in the next section.

*D. Reconstruction of The Simplified Space*

To explore the usefulness of diverse auxiliary tasks for large-scale optimization, instead of using one fixed simplified problem space, the proposal is made to build multiple simplified problem spaces periodically while the evolutionary search progresses online. In particular, if the reconstruction of the simplified problem space occurs in generation $G_2$, the PCA used in section III-A is considered here again to reconstruct a simplified problem space with a new set of solutions in the original problem space. Further, in order to preserve the useful traits found in the last simplified space, the solutions in archive $A_s$ are used as the new set of solutions and subjected to PCA o construct a new $\mathbf{P}_s$[1]. Subsequently, the solutions in $A_s$ and corresponding mapped solutions in $\mathbf{P}_s$ will be used to learn mapping $M_{P_s \to P}$ and $M_{P \to P_s}$ across problem spaces $\mathbf{P}$ and $\mathbf{P}_s$. Finally, the population of the simplified problem space is also re-initialized in the new $\mathbf{P}_s$, which is shown in detail in Alg. 1.

---

[1] In this study, for simplicity, we kept the dimension of $\mathbf{P}_s$ as $d_s$ unchanged.

**Algorithm 2:** Pseudo code of the proposed multi-space evolutionary search for large-scale optimization

**Input:** $\mathbf{P}$: Given the problem space of interest; $d_s$: Dimensionality of the simplified problem space; $G_t$: Interval of knowledge transfer across problem spaces; $G_r$: Interval of simplified space reconstruction

**Output:** $s^*$: optimized solution of the given problem

1: **Begin**
2: **Construct** the simplified problem space $\mathbf{P}_s$ with dimension $d_s$ of given problem $\mathbf{P}$;
3: **Learn** the mappings $M_{P_s \to P}$ and $M_{P \to P_s}$ across $\mathbf{P}_s$ and $\mathbf{P}$;
4: $gen = 1; A_s = \emptyset$
5: **while** Terminate condition is not satisfied **do**
6:    $gen = gen + 1$;
7:    Perform reproduction operators (e.g., crossover and mutation) for $\mathbf{P}_s$ and $\mathbf{P}$, respectively;
8:    Transform solutions in $\mathbf{P}_s$ back to $\mathbf{P}$;
9:    Perform natural selection for both $\mathbf{P}_s$ and $\mathbf{P}$ in problem space $\mathbf{P}$;
10:    **if** $mod(gen, G_t) = 0$ **then**
11:      **Perform** knowledge transfer across $\mathbf{P}_s$ and $\mathbf{P}$;
12:      Transform the population of $\mathbf{P}_s$ back to $\mathbf{P}$, and archive the population in $A_s$;
13:      Remove the repeated solutions in $A_s$;
14:    **end if**
15:    **if** $mod(gen, G_r) = 0$ **then**
16:      **Reconstruct** the new simplified problem space $\mathbf{P}_s$ with dimension $d_s$ using solutions in $A_s$;
17:      Learn the new mappings $M_{P_s \to P}$ and $M_{P \to P_s}$ across the new constructed $\mathbf{P}_s$ and $\mathbf{P}$;
18:      Re-initialize the populations in $\mathbf{P}_s$ using the new learned $M_{P \to P_s}$;
19:    **end if**
20: **end while**
21: **End**

*E. Summary of Proposed Multi-Space Evolutionary Search*

A summary of the proposed multi-space evolutionary search for large-scale optimization is presented in Alg. 2. As can be observed, while the EMT progresses online, the knowledge transfer across $\mathbf{P}_s$ and $\mathbf{P}$ occurs in every in every $G_t$ generation (see line 10-13 in Alg. 2), and the reconstruction of a new simplified problem space $\mathbf{P}_s$ is performed in every $G_r$ generations (see line 15-18 in Alg. 2). Further, archive $A_s$ on line 12-13 of Alg. 2 is used to store the non-repeating search traces in the simplified problems with solution representation in the original problem space. Without loss of generality, the volume of $A_s$ can be configured as needed. However, in this paper, for simplicity, the volume of $A_s$ is configured as $5 * NP$, where NP denotes the population size of the evolutionary search. Once the search traces exceed the volume of $A_s$, only the latest search traces are archived.

## IV. Empirical Study

This section discusses the results of comprehensive empirical studies that were conducted to evaluate the performance of the proposed multi-space evolutionary search paradigm on the commonly used large-scale optimization benchmarks, compared to several state-of-the-art algorithms proposed in the literature.

*A. Experimental Setup*

In this study, the commonly used CEC2013 large-scale optimization benchmark [49], which contains 15 functions with diverse properties, was used to investigate the performance of the proposed multi-space evolutionary search. As summarized in Table I, according to [49], the benchmark consists of both unimodal and multimodal minimization functions, which can be generally categorized into the following five classes: 1) fully separable functions, 2) partially additive separable functions I, 3) partially additive separable functions II, 4) overlapping functions, and 5) fully nonseparable functions. Further, except for functions F13 and F14, all the functions have a dimensionality of 1000. Because of the overlapping property, functions F13 and F14 both have 905 decision variables. For more details on the CEC2013 large-scale optimization benchmark, interested readers can refer to [49].

Table I Properties of the CEC2013 Benchmark Functions

| Separability | Function | Modality | Search Space | Base Function |
|---|---|---|---|---|
| Fully Separable Functions | F1 | unimodal | $[-100,100]^D$ | Elliptic |
| | F2 | multimodal | $[-5,5]^D$ | Rastrigin |
| | F3 | multimodal | $[-32,32]^D$ | Ackley |
| Partially Additive Separable Functions I | F4 | unimodal | $[-100,100]^D$ | Elliptic |
| | F5 | multimodal | $[-5,5]^D$ | Rastrigin |
| | F6 | multimodal | $[-32,32]^D$ | Ackley |
| | F7 | multimodal | $[-100,100]^D$ | Schwefel |
| Partially Additive Separable Functions II | F8 | unimodal | $[-100,100]^D$ | Elliptic |
| | F9 | multimodal | $[-5,5]^D$ | Rastrigin |
| | F10 | multimodal | $[-32,32]^D$ | Ackley |
| | F11 | unimodal | $[-100,100]^D$ | Schwefel |
| Overlapping Functions | F12 | multimodal | $[-100,100]^D$ | Rosenbrock |
| | F13 | unimodal | $[-100,100]^D$ | Schwefel |
| | F14 | unimodal | $[-100,100]^D$ | Schwefel |
| Fully Non-separable Functions | F15 | unimodal | $[-100,100]^D$ | Schwefel |

Next, to verify the efficacy of the proposed multi-space evolutionary search (referred to as MSES hereafter) for large-scale optimization, four state-of-the-art methods for addressing large-scale optimization, including decomposition based cooperative coevolution and non-decomposition-based approaches, were considered as the baseline algorithms for comparison. In particular, the cooperative coevolution approaches included the recursive decomposition method proposed by Sun et al. in 2018 (called RDG) [41], and an improved variant of the differential grouping algorithm introduced by Omidvar et al. which is called DG2 [50]. The non-decomposition-based approaches included the level based learning swarm optimizer proposed by Yang et al. in 2018 (called DLLSO) [51], and the random embedding-based method proposed by Hou et al. in 2019 (called MeMAO) [44]. Further, in these compared algorithms, it should be noted that different evolutionary search methods were used as the basic optimizer. For example, RDG and DG2 employed the self-adaptive differential evolution with neighborhood search (SaNSDE) [41], [50] as the optimizer, while MeMAO considered the classical differential evolution (DE) method as the optimizer [44]. Rather than using differential evolution, DLLSO used the particle swarm optimizer as the basic search method [51]. For a fair comparison to the different baseline algorithms, the optimizer for each space in the proposed MSES was kept consistent with the optimizer used in the compared algorithm. Lastly, the parameter and operator settings of all the compared algorithms and the proposed MSES were kept the same as those in [41], [50], [51], and [44], which are summarized as follows:

- Population Size: population size $NP = 50, 100,$ and 500 for optimizers SaNSDE, DE, and DLLSO, respectively.
- Independent number of runs: $runs = 25$ for all the compared algorithms.
- Maximum number of fitness evaluations: $Max\_FEs = 3E + 06$.
- Number of solutions to be transferred across spaces in MSES: $P = Q = 0.2 * NP$.
- Interval of knowledge transfer across problem spaces: $G_t = 1$.
- Interval of simplified space reconstruction: $G_r = 10$.
- Dimensionality of the simplified problem space: $d_s = 600$.
- Size of $A_s$: $A_{ssize} = 5 * NP$.

B. Results and Discussion

This section presents and discusses the performance of the proposed MSES in comparison to those of the existing state-of-the-art approaches on the CEC2013 large-scale benchmark functions in terms of the solution quality and search efficiency.

*1) Solution Quality*: Table II tabulates the results with respect to the averaged objective values and standard deviations obtained by all the compared algorithms over 25 independent runs. In particular, based on the evolutionary solver employed for the search (e.g., SaNSDE, PSO, and DE), the comparison results are divided into three groups, with each group sharing the same evolutionary solver. The best performance in each comparison group is highlighted using a bold font in the table. Further, in order to obtain a statistical comparison, a Wilcoxon rank sum test with a 95% confidence level was conducted on the experimental results, where "+", "-", and "≈" show that statistically the algorithm is significantly better than, significantly worse than, or similar to the proposed MSES, respectively.

As can be observed in the table, in all three comparison groups, when using different evolutionary search methods as the optimizer, the proposed MSES obtained a superior solution quality in terms of the averaged objective value on most of the problems compared to the other algorithms. In the comparison groups using SaNSDE and PSO as the optimizers, the proposed approach, that is, MSES$_{SaNSDE}$ and MSES$_{DLLSO}$, lost to the compared algorithms on large-scale benchmarks F1 and F2. Table I shows that F1 and F2 are fully separable functions. Moreover, F1 is based on a unimodal "Elliptic" function, and the search space of F2 is only within the range of [-5, 5], which indicates the simplicity of the search spaces for these two functions. However, on the other more complex large-scale benchmarks such as partially additive separable, overlapping, and fully nonseparable problems, where greater appropriate guidance is required for an effective search, the proposed MSES$_{SaNSDE}$ and MSES$_{DLLSO}$ achieved superior and competitive averaged objective values in contrast to DG2/RDG and DLLSO, respectively. On benchmarks F11, F13 and F14, only the proposed method was able to consistently find solutions with objective values of approximately e+07 in both of these comparison groups. On 15 large-scale benchmarks, the proposed MSES$_{SaNSDE}$ and MSES$_{DLLSO}$ achieved significantly better averaged objective values on 13 and 9 problems in contrast to DG2/RDG and DLLSO, respectively.

Furthermore, in the comparison group of that used DE as the optimizer, the proposed MSES$_{DE}$ obtained superior or competitive averaged objective values on all the large-scale benchmarks compared to MeMAO. In particular, on benchmarks such as F4 and F8, MSES$_{DE}$ achieved improvements of orders of magnitude in contrast to MeMAO. The objective values achieved on these benchmarks were even superior to those obtained using SaNSDE and PSO as the optimizers. On 15 large-scale benchmarks, the proposed MSES$_{DE}$ achieved significantly better averaged objective values on 13 problems in contrast to MeMAO.

Table II Averaged objective values and standard deviations obtained by the proposed MSES and the compared baseline algorithms. (Superior performance in each comparison group is highlighted in bold, "+", "≈" and "−" denote that the compared algorithm is statistically significant better, similar and worse than the proposed MSES using different EA solvers, respectively.)

| Problems | Comparison 1 | | | Comparison 2 | | Comparison 3 | |
|---|---|---|---|---|---|---|---|
| | MSES$_{SaNSDE}$ | DG2 | RDG | MSES$_{DLLSO}$ | DLLSO | MSES$_{DE}$ | MeMAO |
| F1 | 1.34e+07±1.54e+06 | 5.20e+02±1.32e+03+ | **2.16e+01±9.09e+01**+ | 1.00e-04±1.51e-04 | **3.98e-22±1.09e-22**+ | **5.03e+07±4.59e+06** | 3.31e+11±2.44e+10− |
| F2 | 1.80e+04±8.60e+02 | **1.26e+04±7.07e+02**+ | 1.28e+04±7.07e+02+ | 2.88e+03±1.96e+02 | **1.12e+03±6.33e+01**+ | **9.09e+03±5.93e+02** | 1.56e+05±7.39e+03− |
| F3 | **2.00e+01±2.39e-03** | 2.14e+01±1.38e-02− | 2.14e+01±1.59e-02− | **2.13e+01±1.54e-01** | 2.16e+01±5.82e-03− | **2.14e+01±2.25e-02** | 2.15e+01±1.69e-02− |
| F4 | **1.29e+09±4.50e+08** | 5.08e+10±1.78e+10− | 4.00e+10±1.22e+10− | **1.30e+09±5.16e+08** | 5.85e+09±1.08e+09− | **1.10e+09±2.58e+08** | 4.70e+12±1.04e+12− |
| F5 | **4.52e+06±7.48e+05** | 5.35e+06±4.84e+05− | 5.05e+06±3.69e+05− | **5.79e+05±6.92e+04** | 6.77e+05±1.16e+05− | **6.33e+06±2.98e+06** | 4.31e+07±4.62e+06− |
| F6 | **1.00e+06±6.14e+03** | 1.06e+06±9.95e+02− | 1.06e+06±1.20e+03− | 1.06e+06±1.60e+03 | **1.06e+06±9.75e+02**≈ | **1.06e+06±9.39e+02** | 1.06e+06±1.59e+03≈ |
| F7 | **1.85e+06±1.67e+05** | 6.55e+07±1.96e+07− | 9.49e+07±5.34e+07− | 1.90e+06±7.26e+05 | **1.58e+06±8.81e+05**≈ | **2.34e+06±3.15e+05** | 5.54e+13±4.98e+13− |
| F8 | **7.14e+12±5.94e+12** | 5.71e+15±1.48e+15− | 4.39e+15±1.76e+15− | **8.73e+11±1.42e+11** | 1.30e+14±4.44e+13− | **1.13e+10±6.25e+09** | 1.56e+17±4.15e+16− |
| F9 | **4.48e+08±7.43e+07** | 4.94e+08±3.05e+07− | 4.98e+08±2.92e+07− | **4.14e+07±7.00e+06** | 4.26e+07±9.49e+06≈ | **1.06e+08±2.20e+07** | 3.25e+09±3.99e+08− |
| F10 | **9.10e+07±5.42e+05** | 9.46e+07±2.78e+05− | 9.45e+07±3.49e+05− | **9.40e+07±2.29e+05** | 9.40e+07±2.61e+05≈ | 9.41e+07±1.52e+05 | **9.40e+07±2.18e+05**≈ |
| F11 | **1.51e+07±4.35e+06** | 4.82e+09±5.60e+09− | 6.17e+08±1.36e+08− | **1.62e+07±1.94e+07** | 2.35e+08±6.58e+07− | **7.95e+07±2.66e+07** | 1.22e+16±7.07e+15− |
| F12 | **1.68e+03±1.80e+02** | 2.62e+05±1.17e+06− | 4.38e+03±1.09e+03− | **1.62e+03±1.84e+02** | 1.77e+03±1.30e+02− | **1.78e+03±2.66e+02** | 1.75e+12±3.55e+11− |
| F13 | **1.51e+07±2.38e+06** | 1.62e+09±4.58e+08− | 3.02e+09±8.26e+08− | **1.11e+07±7.44e+06** | 3.26e+08±1.22e+08− | **3.06e+07±6.95e+06** | 1.10e+16±7.24e+15− |
| F14 | **4.83e+07±7.52e+06** | 5.26e+09±2.86e+09− | 3.73e+09±1.95e+09− | **2.48e+07±8.72e+06** | 2.03e+08±2.41e+08− | **6.52e+07±8.96e+06** | 1.63e+16±9.69e+15− |
| F15 | **1.64e+06±1.21e+05** | 1.11e+07±1.73e+06− | 9.76e+06±1.41e+06− | **1.49e+06±1.55e+05** | 4.51e+06±4.70e+05− | **3.59e+06±3.21e+05** | 1.32e+18±5.28e+17− |
| +/≈/- No. | - | 2/0/13 | 2/0/13 | - | 2/4/9 | - | 0/2/13 |

In summary, because the proposed method used the same optimizer as the compared algorithms in each comparison group, and only differed in the search spaces, the superior solution quality observed in Table II confirmed the effectiveness of the proposed multi-spaces evolutionary search for large-scale optimization.

*2) Search Efficiency*: This section presents the convergence graphs obtained by all the compared algorithms on all the large-scale benchmarks to assess the search efficiency of the proposed multi-spaces evolutionary search for large-scale optimization. In particular, Fig. 3 and Fig. 4 show the obtained convergence graphs obtained on the fully separable functions, partially additive separable functions, overlapping functions, and fully non-separable functions. In these figures, the Y-axis denotes the averaged objective values in log scale, while the X-axis gives the respective computational effort required in terms of the number of fitness evaluations.

It can be observed from Fig. 3 and Fig. 4 that on benchmarks F1 and F2, the compared algorithm DLLSO obtained the best convergence performance. This indicates that on benchmarks with a relatively simple decision space, a search on the original problem space can efficiently find high-quality solutions using properly designed search strategies.

Moreover, on the other functions of the CEC 2013 benchmarks with more complex decision spaces (e.g., partially additive separable functions, overlapping functions, and fully nonseparable functions), where greater appropriate guidance is required for an efficient search for high-quality solutions, the proposed MSES obtained the best and competitive convergence performances in all three comparison groups. In particular, even on the fully separable function F3, because it is based on the complex "Ackley" function, the proposed $MSES_{SaNSDE}$, $MSES_{DLLSO}$, and $MSES_{DE}$ obtained faster convergences over the compared algorithms that shared the same EA solvers. In addition, for functions such as F8, F13, and F14, regardless of which EA was considered as the search optimizer, the proposed MSES obtained the best convergence performance in contrast to the baseline algorithms in all three comparison groups. Because the proposed MSES used the same search optimizers as the compared algorithms, the superior search speed obtained confirmed the efficiency of the proposed multispace evolutionary search for large-scale optimization.

Finally, to provide deeper insights into the superior performance obtained by the proposed MSES, considering the three different search algorithms (SaNSDE, PSO, and DE) as the optimizers, the transferred solutions from the simplified space and the best solutions in the population of the original problem space on the representative benchmarks are plotted in Fig. 5. As can be observed in the figure, solutions were transferred across the problem spaces during the evolutionary search process. In particular, in Fig. 5(a), compared to the best solution in the original problem space at different stages of the search, both inferior and superior solutions in terms of the objective value were transferred across the spaces. The former could be eliminated via natural selection, while the latter survived and efficiently guided the evolutionary search in the original problem space toward promising areas of high-quality solutions, which led to the enhanced search performance of the proposed MSES, as observed in Table II, Fig. 3, and Fig. 4. Similar observations can also be made in the cases of using PSO and DE as the optimizers, as depicted in Fig. 5(b) and Fig. 5(c), respectively. These also confirmed that useful traits could be embedded in the different spaces of a given problem, and concurrently conducting an evolutionary search on multiple spaces can lead to efficient and effective problem-solving for large-scale optimization.

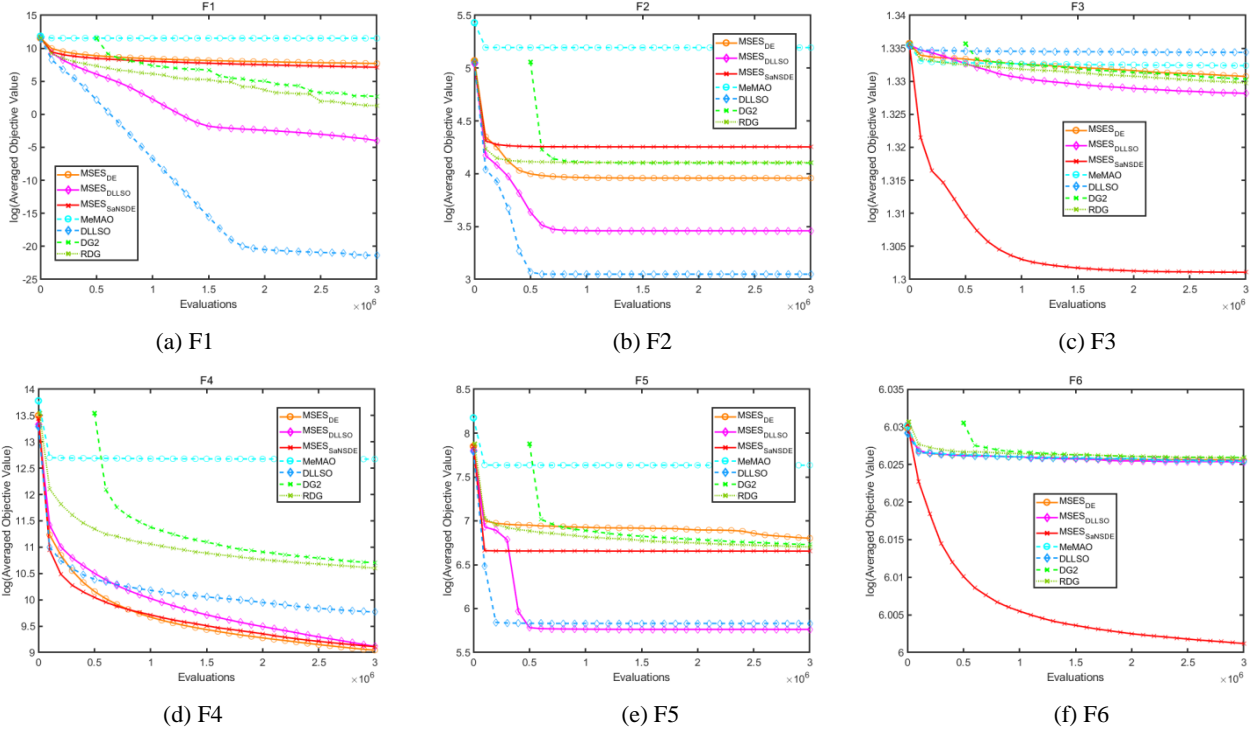

Fig. 3. Convergence curves of average fitness (over 25 independent runs) obtained by MSES and the compared algorithms on CEC2013 fully separable and partially additive separable functions, i.e., F1-F6. Y-axis: averaged objective value in log scale; X-axis: number of fitness evaluations.

*3) Sensitivity Study*: Five parameters were used in the proposed multi-space evolutionary search: the size of $A_s$ ($A_{ssize}$), dimension of the simplified problem space ($d_s$), interval for reconstructing the simplified problem space ($G_r$), and interval and number of solutions transferred from the simplified to the original problem space ($G_t$ and $Q$, respectively). This section presents and discusses how these parameters affect the performance of the proposed multi-space evolutionary search.

In particular, Figs. 6 to 8 present the averaged objective values obtained by the proposed MSES and DLLSO on the representative benchmarks across 25 independent runs with different configurations of $A_{ssize}$, $d_s$, $Q$, $G_t$ and $G_r$. In the figures, the X-axis gives different benchmark functions, while the Y-axis denotes the normalized averaged objective value obtained by each compared configuration. Specifically, the obtained objective values on each benchmark are normalized by the worst (largest) objective obtained by all the compared algorithms on the benchmark. Therefore, values close to 0 and 1 denote the best and worst performances, respectively. Further, because DLLSO was observed to obtain a superior solution quality and search speed in contrast to the other compared algorithms in sections IV-B1 and IV-B2, it is considered as the baseline algorithm here. For a fair investigation, DLLSO was also used as the optimizer in the proposed MSES.

Out of these parameters, $A_{ssize}$, $d_s$, and $G_r$ were involved in the simplified problem space construction. $A_{ssize}$ defined the number of solutions for constructing the simplified problem space, while $d_s$ gave the dimensionality of the constructed space. Further, $G_r$ determined the frequency for reconstructing the simplified problem space. Generally, small numbers of $A_{ssize}$ and $d_s$ simplified the problem space to a large extent, and large numbers of these two parameters could make the constructed space close to the original problem space. In addition, small and large values for $G_r$ reconstructed the low-dimensional space frequently and infrequently during the evolutionary search, respectively. As can be observed in Fig. 6, on the partially additive separable functions, for example, F5 and F8, *NP* (population size) number of solutions are already able to construct

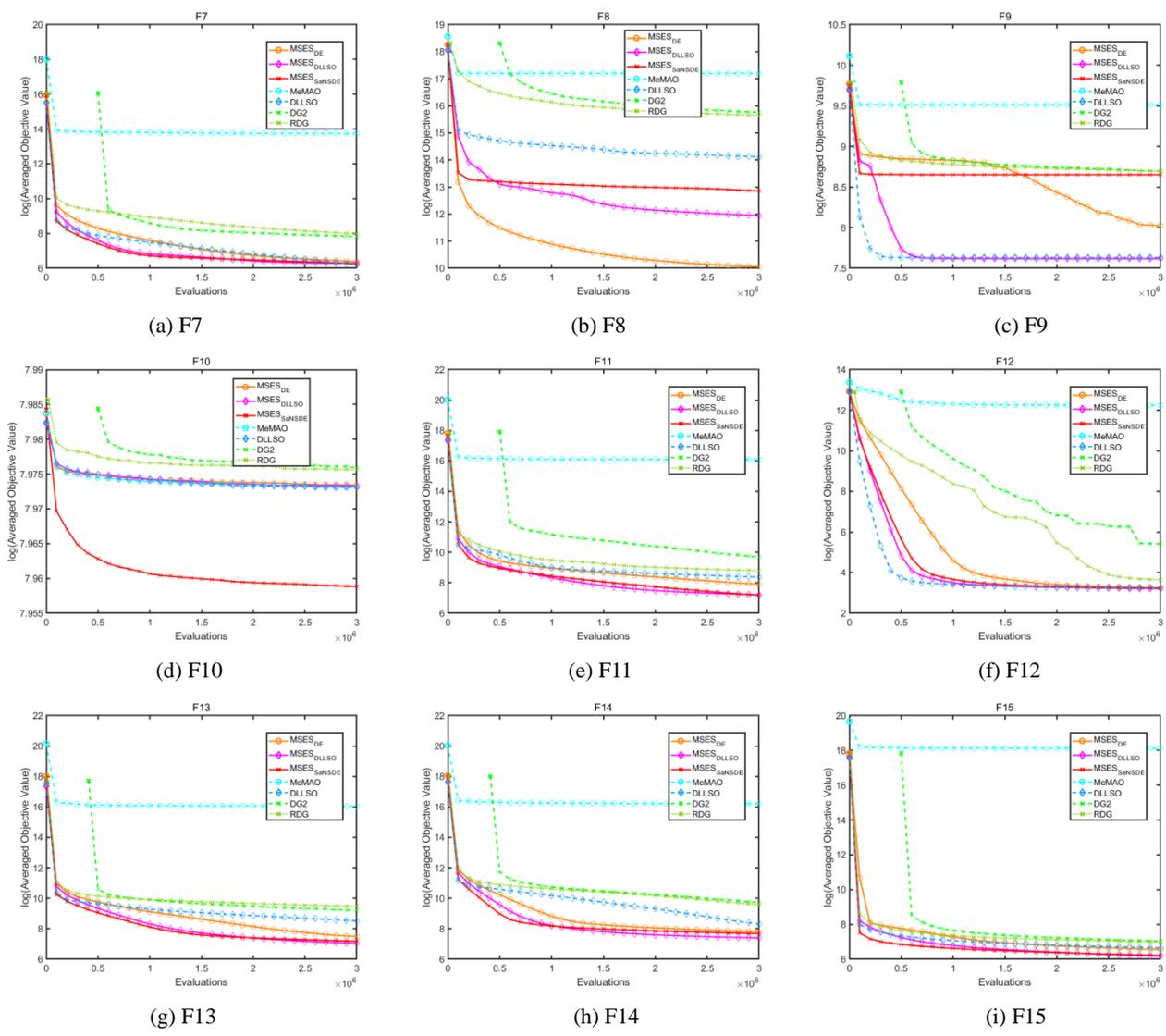

Fig. 4. Convergence curves of average fitness (over 25 independent runs) obtained by MSES and the compared algorithms on CEC2013 partially additive separable, overlapping and fully non-separable functions, i.e., F7-F15. Y-axis: averaged objective value in log scale; X-axis: number of fitness evaluations.

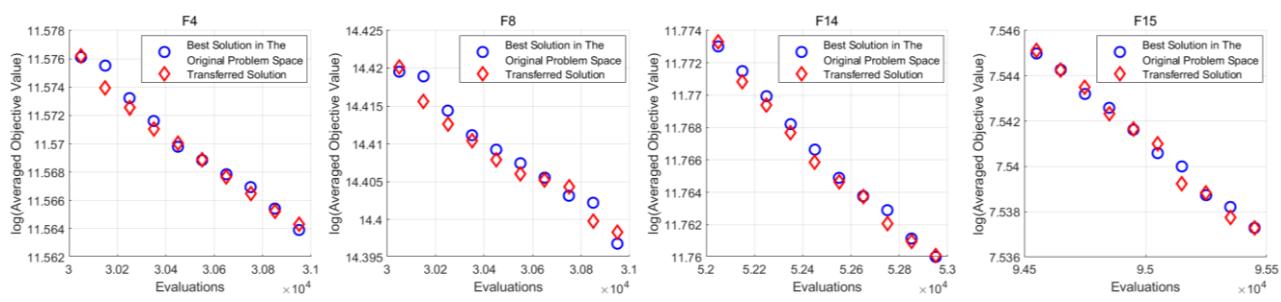

(a) Comparison 1: SaNSDE as the EA solver

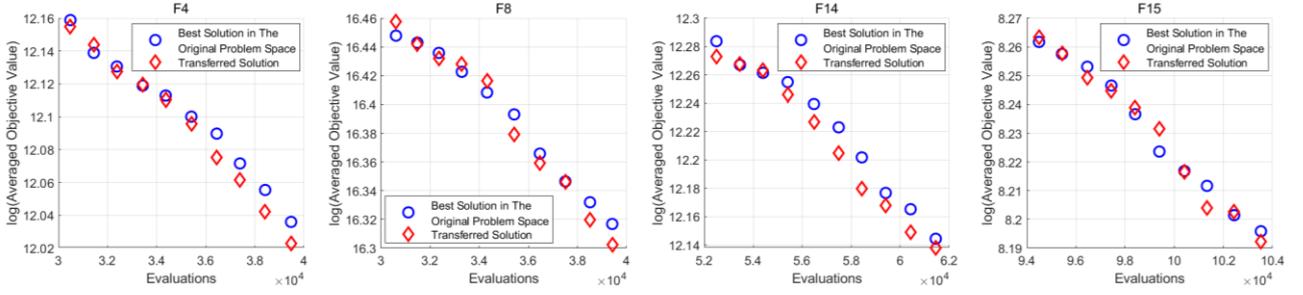

(b) Comparison 2: PSO as the EA solver

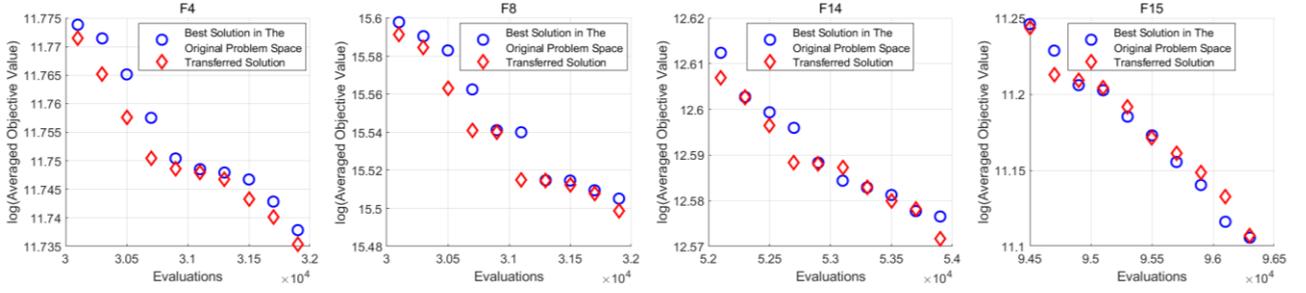

(c) Comparison 3: DE as the EA solver

Fig. 5. Illustration of the transferred solutions and the best solution in the population on representative benchmarks of different comparison groups.

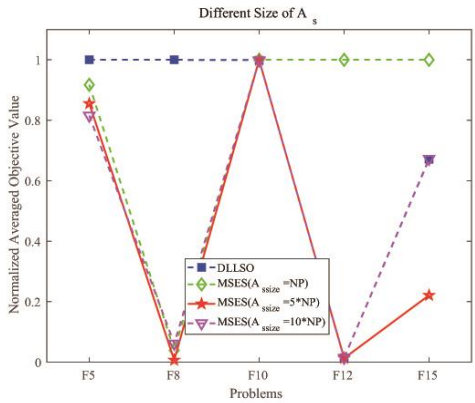

Fig. 6. Averaged objective values obtained by the proposed MSES and DLLSO on representative benchmarks across 25 independent runs with various configurations of $A_{ssize}$.

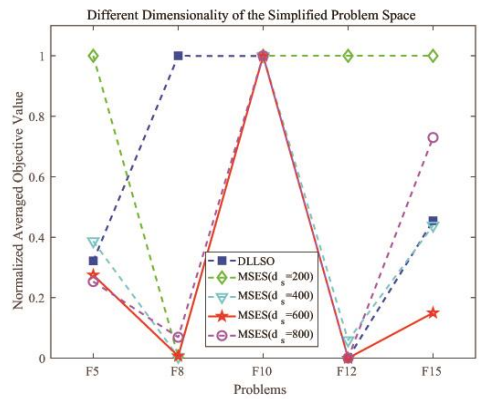

Fig. 7. Averaged objective values obtained by the proposed MSES and DLLSO on representative benchmarks across 25 independent runs with various configurations of $d_s$.

a useful problem space that can improve the search in the original problem space (see the superior objective values achieved by MSES with $A_{ssize} = NP$, $A_{ssize} = 5*NP$, and $A_{ssize} = 10*NP$). However, on the more complex functions, for example, F12 and F15, a larger $A_{ssize}$ may be required to provide more information for constructing a useful problem space in MSES. Furthermore, for the dimensionality of the simplified problem space, as can be observed in Fig. 7, neither a small nor a large number for $d_s$ is good for building a useful simplified problem space because a very low dimensionality could lose important information for efficient evolutionary search, and a space with a dimensionality close to the original problem cannot play a complementary role to the original problem space for the proposed MSES. Lastly, as depicted in Fig. 8, the frequency of reconstructing the simplified space did not significantly affect the performance of MSES on the considered large-scale benchmarks.

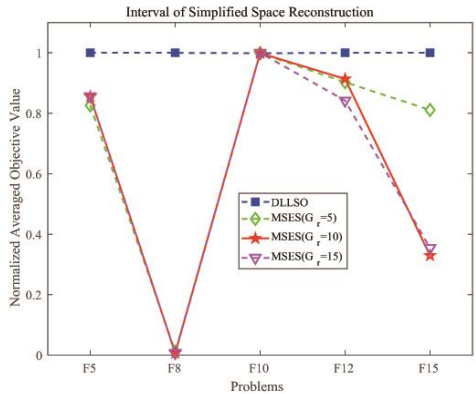

Fig. 8. Averaged objective values obtained by the proposed MSES and DLLSO on representative benchmarks across 25 independent runs with various configurations of $G_r$.

On the other hand, parameters $Q$ and $G_t$ defined the amount and frequency of knowledge sharing across problem spaces. Generally, a small value of $Q$ and large value of $G_t$ significantly reduced the amount and frequency of solution transfer across spaces, while a large value of $Q$ small value of $G_t$ greatly increased the amount and frequency of knowledge sharing across problem spaces. It can be observed from Fig. 9 and Fig. 10, with different configurations of $Q$ and $G_t$ values, superior solution qualities were obtained by the proposed MSES compared to DLLSO on most of the benchmarks. However, while the optimal confirmations of these parameters were generally problem-dependent, the configuration considered in the empirical study, as discussed above, was found to provide noteworthy results across a variety of larger-scale optimization problems.

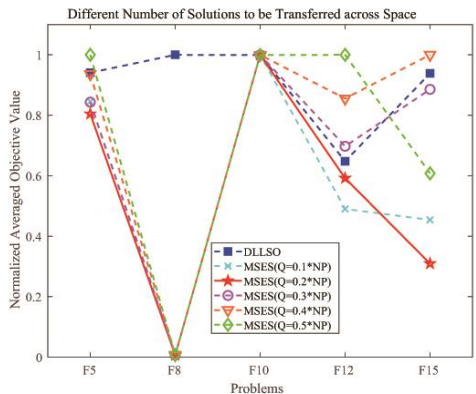

Fig. 9. Averaged objective values obtained by the proposed MSES and DLLSO on representative benchmarks across 25 independent runs with various configurations of $Q$

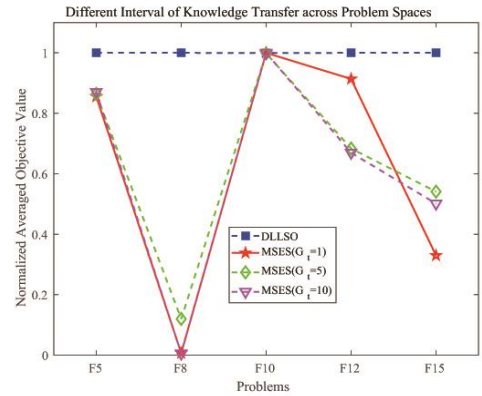

Fig. 10. Averaged objective values obtained by the proposed MSES and DLLSO on representative benchmarks across 25 independent runs with various configurations of $G_t$.

## V. Conclusions

This paper proposed a multi-space evolutionary search paradigm for large-scale optimization. In particular, it presented the details of the problem space construction, learning of the mapping across problem spaces, and knowledge transfer across problem spaces. In contrast to existing methods, the proposed paradigm conducts an evolutionary search on multiple solution spaces derived from the given problem, each possessing a unique landscape. More importantly, the proposed paradigm makes no assumptions about the given large-scale optimization problem, such as that the problem is decomposable or that a certain relationship exists among the decision variables. To validate the performance of the proposed paradigm, comprehensive empirical studies on the CEC2013 large-scale benchmark problems were conducted. The results were compared to those of recently proposed largescale evolutionary algorithms, which confirmed the efficacy of the proposed multi-space evolutionary search for large-scale optimization.

Future work will further explore effective approaches for constructing simplified problem spaces in the proposed multi-space evolutionary search for efficient problem solving in large-scale optimization. The design of adaptive parameter configurations in the proposed paradigm is also a promising research direction for improving the generality of a multi-space evolutionary search for large-scale optimization.